\documentclass[review]{elsarticle}
\graphicspath{ {./figures/} }
\usepackage{hyperref}
\usepackage{float}
\usepackage{verbatim} 
\usepackage{apalike}
\restylefloat{figure}
\floatstyle{plaintop} 
\restylefloat{table}
\usepackage{xcolor}

\usepackage{amsmath}
\usepackage{amssymb}
\usepackage{tikz}
\usepackage{multirow}
\usepackage{graphicx}
\usepackage[caption=false,font=normalsize,labelfont=sf,textfont=sf]{subfig}
\usepackage{caption}
\usepackage{subcaption}

\usepackage{float}
\journal{Expert Systems with Applications}

\biboptions{authoryear}

\begin{document}
	\begin{frontmatter}

		\title{High-Frequency Enhanced Hybrid Neural Representation for Video Compression}
		
		\author[label1,label2]{Li Yu \corref{cor1}}
		\ead{li.yu@nuist.edu.cn}
		
		\author[label1]{Zhihui Li}
		\ead{202212490325@nuist.edu.cn}
		
		\author[label3]{Jimin Xiao}
		\ead{jimin.xiao@xjtlu.edu.cn}
  
  	     \author[label4]{Moncef Gabbouj}
		\ead{moncef.gabbouj@tuni.fi}
		
		\cortext[cor1]{Corresponding author.}
		\address[label1]{School of Computer Science, Nanjing University of Information Science and Technology, Nanjing, 210044, China}
		\address[label2]{Jiangsu Collaborative Innovation Center of Atmospheric Environment and Equipment Technology, Nanjing University of Information Science and Technology, Nanjing, 210044, China}
		\address[label3]{Department of Intelligent Science, School of Advanced Technology, Xi’an Jiaotong-Liverpool University, Suzhou, 215000, China}
  	\address[label4]{Faculty of Information Technology and Communication Sciences, Tampere University, Tampere, 33101, Finland}
		
		\begin{abstract}
			Neural Representations for Videos (NeRV) have simplified the video codec process and achieved swift decoding speeds by encoding video content into a neural network, presenting a promising solution for video compression. However, existing work overlooks the crucial issue that videos reconstructed by these methods lack high-frequency details. To address this problem, this paper introduces a High-Frequency Enhanced Hybrid Neural Representation Network. Our method focuses on leveraging high-frequency information to improve the synthesis of fine details by the network. Specifically, we design a wavelet high-frequency encoder that incorporates Wavelet Frequency Decomposer (WFD) blocks to generate high-frequency feature embeddings. Next, we design the High-Frequency Feature Modulation (HFM) block, which leverages the extracted high-frequency embeddings to enhance the fitting process of the decoder. Finally, with the refined Harmonic decoder block and a Dynamic Weighted Frequency Loss, we further reduce the potential loss of high-frequency information. Experiments on the Bunny and UVG datasets demonstrate that our method outperforms other methods, showing notable improvements in detail preservation and compression performance.
		\end{abstract}
		
		\begin{keyword}
			 Neural representation for videos\sep video compression\sep high-frequency information\sep wavelet transform
		\end{keyword}
		
	\end{frontmatter}
	
	\section{Introduction}
	\label{introduction}
	
Nowadays, with the proliferation of high-speed Internet and streaming services, video content has become the dominant component of Internet traffic. According to statistics, in 2023, more than 65\% of total Internet traffic is video content \citep{video2023}, and this percentage is expected to continue increasing. In the past, video compression was usually achieved by traditional codecs like H.264/AVC \citep{h264}, H.265/HEVC \citep{h265}, H.266/VVC \citep{h266}, and AVS \citep{AVS}. However, the handcrafted algorithms in these traditional codecs would limit the compression efficiency. 

With the rise of deep learning, many neural video codec (NVC) technologies have been proposed \citep{dvc,dcvc,ssf,temporal}. These approaches replace handcrafted components with deep learning modules, achieving impressive rate-distortion performance. However, these NVC approaches have not yet achieved widespread adoption in practical applications. One reason for this is that these approaches often require a large network to achieve generalized compression over the entire data distribution, which is more computationally intensive and frequently leads to slower decoding speeds compared to traditional codecs. Moreover, the generalization capability of the network depends on the dataset used for model training, leading to poor performance on out-of-distribution (OOD) data from different domains \citep{out}, and even when the resolution changes.

To overcome these challenges associated with NVCs, researchers have turned to implicit neural representations (INRs) as a promising alternative. Recently, INRs have gained increasing attention for their ability to efficiently represent and encode diverse scenes \citep{nerf,3d1}, images \citep{coin,coin++,imgc}, and videos \citep{ipf,mvc,nerv}. Given their simplicity and efficiency, several studies have suggested their application to video compression tasks. In this paradigm, the video is encoded into the parameters of the neural network and can be recovered through the forward pass process of the network. Then, standard neural network compression methods can be applied to further reduce the size of network. This innovative paradigm greatly simplifies the codec process and achieves higher decoding speeds compared to traditional codecs and NVC. Moreover, this method avoids domain generalization issues by tailoring a neural network specifically to each individual video. Specifically, NeRV \citep{nerv} proposes a novel neural representation method based on Convolutional Neural Networks (CNNs). Unlike previous work \citep{coin,coin++}, which mapped coordinates to color values of each pixel, NeRV directly learns the mapping from frame indexes to video frames, greatly improving both coding speed and reconstruction quality. Later, HNeRV \citep{hnerv} proposed a hybrid method that encodes each frame into a corresponding embedding, providing content prior to the decoder for reconstruction. The hybrid-based methods further enhance reconstruction quality and convergence speed, providing the ability to reconstruct high-resolution and high-fidelity video frames.
\begin{figure}[htb]
    \centering
    \begin{minipage}[c]{0.32\linewidth}
        \centering
        \includegraphics[width=\linewidth]{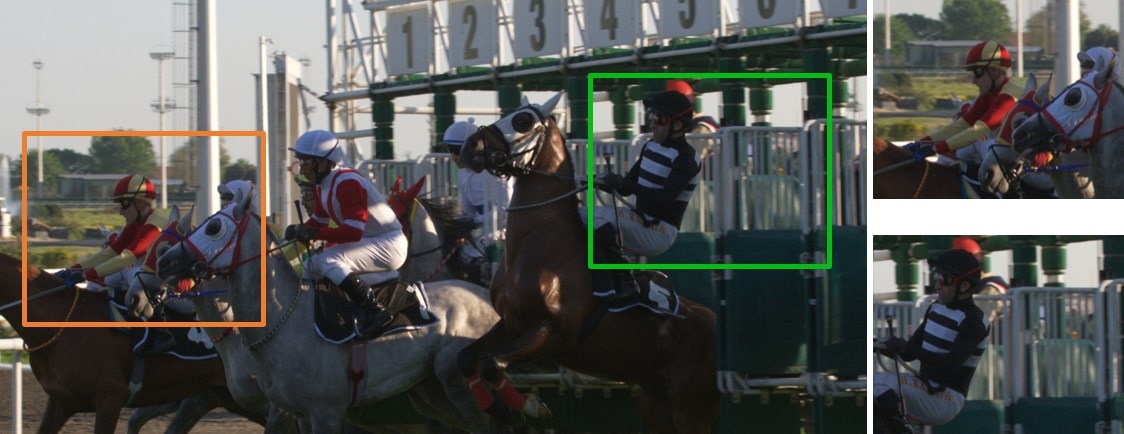}
    \end{minipage}
    \begin{minipage}[c]{0.32\linewidth}
        \centering
        \includegraphics[width=\linewidth]{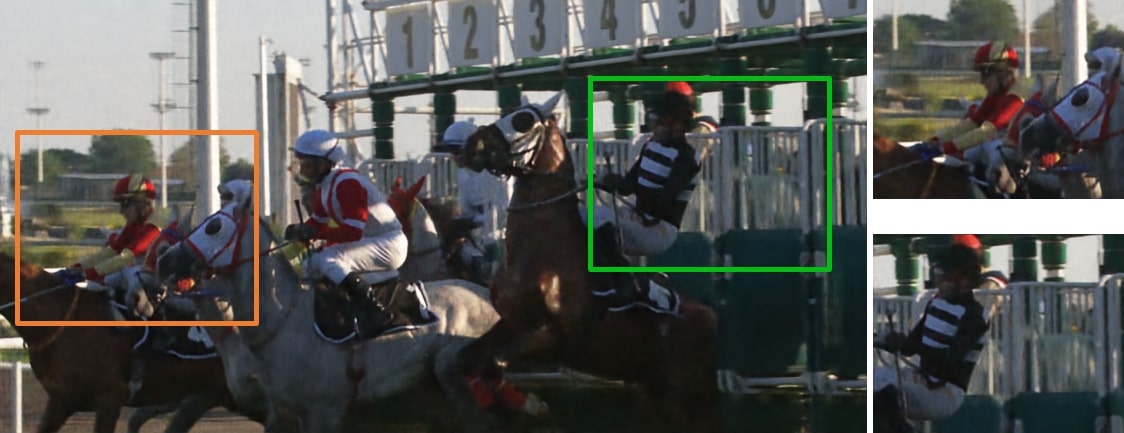}
    \end{minipage}
    \begin{minipage}[c]{0.32\linewidth}
        \centering
        \includegraphics[width=\linewidth]{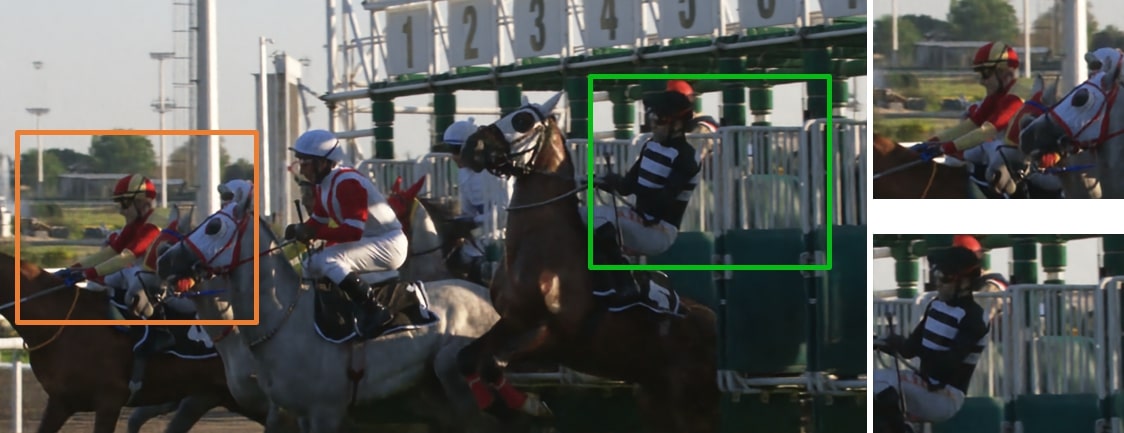}
    \end{minipage}
    
    \vspace{3pt}
    
    \begin{minipage}[c]{0.32\linewidth}
        \centering
        \includegraphics[width=\linewidth]{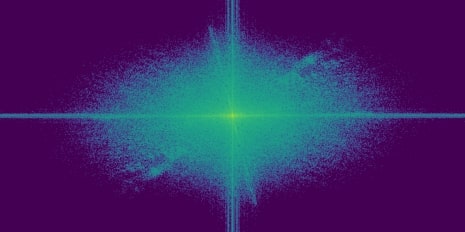}
        \parbox[t]{\linewidth}{\centering \footnotesize Original}
    \end{minipage}
    \begin{minipage}[c]{0.32\linewidth}
        \centering
        \includegraphics[width=\linewidth]{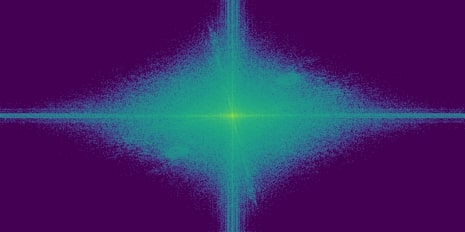}
        \parbox[t]{\linewidth}{\centering \footnotesize HNeRV}
    \end{minipage}
    \begin{minipage}[c]{0.32\linewidth}
        \centering
        \includegraphics[width=\linewidth]{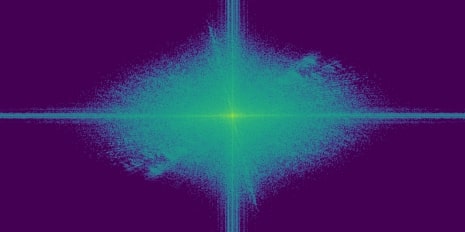}
        \parbox[t]{\linewidth}{\centering \footnotesize Ours}
    \end{minipage}
\caption{Frames and their corresponding frequency maps. Row 1: Original frame alongside frames reconstructed by the HNeRV and our proposed method. Row 2: Frequency maps of the corresponding reconstructed frames, where frequency increases outward from the center. Our method shows better spatial and spectral alignment with the original frame.}
    \label{data:freq_map}
\end{figure}

Although such methods have made impressive progress, the spectral bias \citep{spectralbias} of neural networks and the loss function defined in the spatial domain hinder networks' ability to accurately fit the input. This results in reconstructed videos that are excessively smooth and lacking texture detail, as shown in Fig. \ref{data:freq_map}. To overcome these limitations, we focus on leveraging high-frequency information and propose a High-Frequency Enhanced Hybrid Neural Representation Network. Concretely, to alleviate the burden on the network of synthesizing high-frequency signals, we introduce an additional encoder branch called the wavelet high-frequency encoder, which incorporates Wavelet Frequency Decomposer (WFD) blocks for frequency selection. The wavelet high-frequency encoder extracts extra high-frequency embeddings that allow the decoder to perceive high-frequency detail features and enhance the fitting process. The WFD block utilizes Haar wavelet transform to efficiently perform frequency separation on intermediate features and encodes them as high-frequency embeddings. Moreover, to effectively utilize high-frequency embeddings for accurate frame reconstruction, we propose a High-Frequency Feature Modulation (HFM) block. This block uses modulation vectors learned from high-frequency information to modulate intermediate features, enhancing both the high-frequency fitting ability and feature fusion capability. Additionally, we introduce a Harmonic block by replacing the Gaussian Error Linear Unit (GELU) activation function in the decoder block with an adaptive harmonic activation function. This introduces a periodic inductive bias into the network, boosting its capability to capture periodic structures and complex textures. Finally, by employing a Dynamic Weighted Frequency Loss, our approach significantly improves the reconstruction of high-frequency details in frames.
The primary contributions of this paper can be summarized as:
\begin{itemize}
    \item 
    We propose a High-Frequency Enhanced Hybrid Neural Representation Network that significantly improves the network's ability to synthesize fine details in video reconstruction. To achieve this, the proposed method constructs a complete pipeline, including high-frequency embedding extraction at the encoder, fusion at the decoder, and a Dynamic Weighted Frequency Loss for supervised training.
    \item 
    The WFD block utilizes a Haar wavelet transform to decompose spatial features into high-frequency and low-frequency signals for frequency selection. Additionally, our proposed HFM block effectively leverages the extracted high-frequency embedding to enhance the model fitting process. By integrating the advantages of frequency decomposition and targeted enhancement, our approach optimizes spatial feature representation and model accuracy.
    \item 
    We introduce a Dynamic Weighted Frequency Loss and refined Harmonic decoder block to further enhance the ability of the network to retain the details in reconstructed videos.
    \item 
    We conduct extensive experiments on Bunny and UVG datasets, and both qualitative and quantitative results demonstrate the effectiveness of our method.
\end{itemize}
\section{Related work}
    \subsection{Implicit Neural Representation for Videos}
    Implicit neural representations (INRs) is an innovative approach, utilizing a neural network to learn a mapping function \(f_{\theta}(x) \rightarrow y\) to map the input coordinates \(x\) to corresponding values \(y\) (e.g., RGB colors). The network is trained to approximate a target function \(g\) such that \(\| f_{\theta}(x) - g(x) \| \leq \epsilon\). Once training is complete, we can obtain a lossy reconstruction of the original signal by performing a forward pass through the trained network. Subsequently, we can apply standard neural network compression methods to further reduce the model's size. Recently, INRs have been increasingly applied in video representation \citep{boosting} because of their potential in applications such as video compression, inpainting, and denoising. Early methods \citep{ipf,pixel-wise2} employed pixel-wise representations that map pixel coordinates to pixel values, resulting in slow decoding speed and poor performance. Later, NeRV \citep{nerv} proposed a frame-wise representation method with convolutional layers, yielding better reconstruction performance and faster training speed than previous pixel-wise methods. NeRV successfully demonstrated that frame-by-frame neural representation is an effective method capable of achieving compression performance comparable to standard video codecs. Following NeRV, subsequent research \citep{E-nerv,ps-nerv,ffnerv,hnerv,dnerv} explored further refinements of network structures. Specifically, E-NeRV \citep{E-nerv} reduces redundant model parameters while retaining representational capability by decomposing the implicit neural representation of images into separate spatial and temporal contexts. 
    HNeRV \citep{hnerv} introduced a hybrid neural representation scheme, replacing the time coordinate \(t\) with content-related compact embeddings as network input, to provide a visual prior and greatly improve reconstruction quality and convergence speed. 
    However, all the above methods primarily focused on improving the network architecture to enhance video regression performance. They overlooked the fact that the reconstructed videos lack detailed textures, leading to overly smooth and blurry results that are unfavorable to human perception.
  
    \subsection{Spectral Bias}
   Several studies have demonstrated that neural networks exhibit spectral bias \citep{spectralbias,frequencyprinciple}, meaning they typically prioritize fitting low-frequency signals and tend to overlook high-frequency signals. These high-frequency signals encode fine image details like vertical and horizontal edges. Missing these high-frequency components can lead the network to synthesize blurry images with undesirable artifacts. To address this bias, several strategies have been proposed for pixel-based methods \citep{fourier,improvedfourier}. For example, \cite{fourier} introduced positional encoding into the input coordinates of MLPs to enhance the ability of INRs to express high-frequency information in signals. However, this positional encoding scheme is only suitable for networks that directly take spatial coordinates or specific feature inputs, limiting its general applicability. In addition, research has focused on the selection of diverse nonlinear activation functions, including sinusoidal functions \citep{siren}, Gaussian functions \citep{gaussian}, and continuous complex Gabor wavelet functions \citep{wire}. These functions result in significantly higher representation accuracy than the Rectified Linear Unit (ReLU) activation function. However, INRs such as the Sinusoidal Representation Network (SIREN) \citep{siren} are heavily reliant on careful initialization and hyperparameter tuning for accurate representation. Moreover, these methods require strong prior knowledge of the solution, resulting in their limited applicability.
   
    \subsection{Wavelet Transform}
    Frequency domain analysis plays a pivotal role in image processing by transforming images into the frequency domain, where different frequency components can be intuitively handled, significantly enhancing processing effectiveness. With the rapid advancement of deep learning, frequency domain analysis provides a fresh perspective for image processing, emerging as an effective tool for improving performance across various computer vision tasks \citep{wavegan, wavefill, fregan, deblurring, multi,Dim2clear,irsam}. The wavelet transform has proven particularly effective in decomposing signals into different frequency components, making it an invaluable tool for image analysis. Researchers have successfully integrated this technique with convolutional neural networks (CNNs) to achieve impressive results. For example, \cite{wavegan} combines the wavelet transform with generative adversarial networks (GANs), enhancing the quality of synthesized images from a frequency domain perspective and applying it to few-shot image generation. Similarly, \cite{wavefill} decomposes images into multiple frequency components and uses these signals to fill in damaged image areas, effectively achieving image repair and reconstruction. \cite{fregan} enhances the frequency awareness of the model by directly decomposing the intermediate features of generators and discriminators into the wavelet domain for frequency domain supervision.  Additionally, \cite{irsam} extracts high-frequency features from infrared images via wavelet transform and utilizes these features to enhance target detection performance under low-contrast and noisy conditions. The wavelet transform offers an effective method for extracting and utilizing high-frequency information in images. Based on this, we have constructed a wavelet high-frequency encoder to harness high-frequency details from images.

\section{Method}
    \subsection{Overview}
\begin{figure*}[htb]
        	\centering
                \includegraphics[width=1\linewidth]{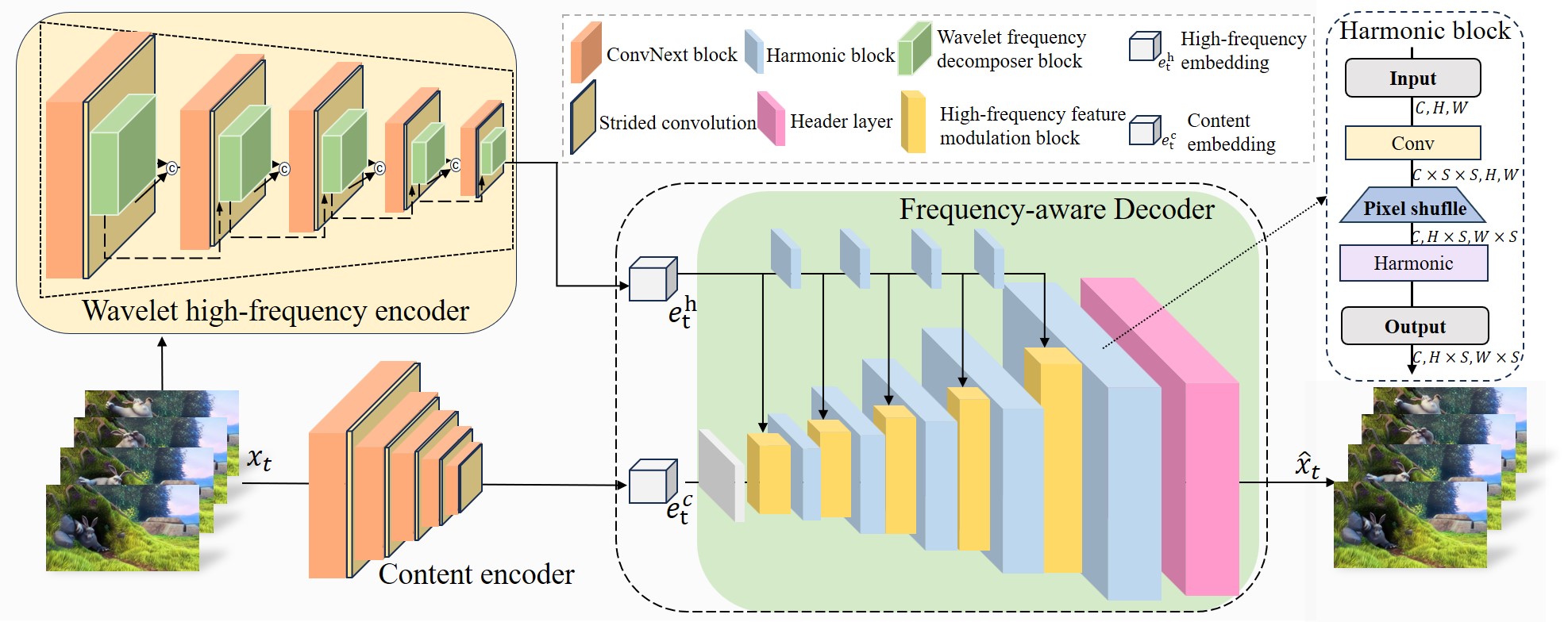}
        	\caption{Overview of our proposed High-Frequency Enhanced Hybrid Neural Representation Network. We encode the content embedding \(e_t^c\) and high-frequency embedding \(e_t^h\) via content encoder and wavelet high-frequency encoder, respectively. On the decoder, we use the proposed Harmonic block to upsample embeddings and use HFM block to fuse \(e_t^c\) and \(e_t^h\). In the end, a header layer converts the features into the final reconstructed image \(\hat{x}_t\). The decoder and embeddings serve as the neural representation of a given video.
         }
     \label{data:model}
        \end{figure*}

    The proposed approach follows the hybrid-based representation method \citep{hnerv} that expresses a video as frame-specific embeddings and a video-specific decoder. Our goal is to train the neural network to overfit the input video to the network parameters, thereby obtaining a neural representation that enables the reconstruction of the video through the embeddings and decoder after training.
      As shown in Fig. \ref{data:model}, our representation network includes three primary components: a content encoder, a wavelet high-frequency encoder, and a frequency-aware decoder. The lightweight ConvNeXt block \citep{convnet} and strided convolution layer are used as the basic encoding unit in two encoder branches. The wavelet high-frequency encoder realizes the filtering and extraction of high-frequency information by integrating the Wavelet Frequency Decomposer (WFD) block. The frequency-aware decoder incorporates the proposed Harmonic block and High-Frequency Feature Modulation (HFM) block to fuse features and perform spatial upsampling.
        
         Let \(\mathcal{X} = \{x_{1},x_{2},x_{3},\ldots,x_{T}\}\) denote the input video sequence, where 
        \(\boldsymbol{x}_{t}\in\mathbb{R}^{H\times W\times3}\) is a video frame at time \(t\) with spatial resolution \(H \times W\).
        The input frame \({x}_{t}\) is initially processed by the content encoder and converted into compact content \({e}_t^{c}\in\mathbb{R}^{\frac{H}{f}\times \frac{W}{f} \times {d_c}}\), which downsamples the spatial dimensions by a factor of \(f\). Meanwhile, the wavelet high-frequency encoder, equipped with WFD block, maps \({x}_{t}\) into high-frequency embedding \({e}_t^{h}\in\mathbb{R}^{\frac{H}{g}\times \frac{W}{g}\times {d_h}}\) with a downsampling factor of \(g\). 
            Subsequently, the obtained \(e_t^h\) is then passed into the HFM block to learn the spatial modulation vector that modulates \(e_t^c\) and then it passes through the proposed Harmonic block to perform spatial upsample. The final reconstructed frame \(\hat{x}_t\) is obtained by the last head layer, which maps the features into the pixel domain.
        Once training is completed, the overfitted model of the input video sequence is then pruned, followed by quantization and entropy coding to generate the final compressed video bitstream. Note that the encoder is discarded after training and is not included in the video bitstream.

    \subsection{Wavelet High-Frequency Encoder}
\begin{figure*}[htb]
    \centering
    
    \includegraphics[width=0.64\linewidth]{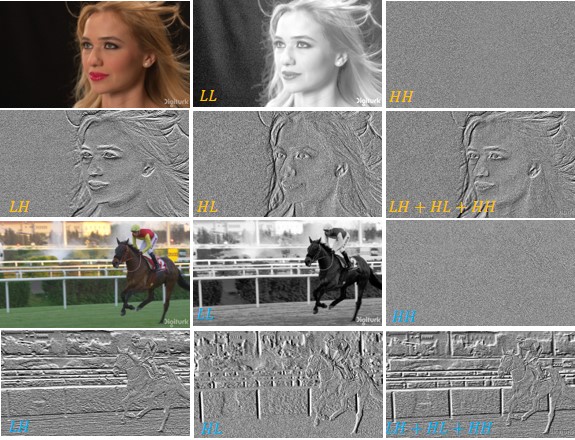}
    \tikz{\draw[dashed] (0,0) -- (0,-6.0);} 
    \includegraphics[width=0.31\linewidth]{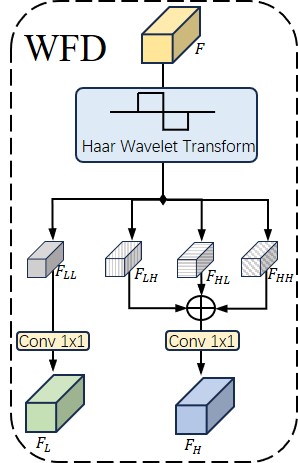}
    
    \caption{(Left) Illustration of frequency components obtained from Haar wavelet transformation. The Haar wavelet transform decomposes the input feature \(F\) into four sub-bands: \(F_{LL}\), \(F_{LH}\), \(F_{HL}\), and \(F_{HH}\). Each transformation reduces the spatial resolution by half. (Right) Illustration of Wavelet Frequency Decomposer block.}
    \label{data:dwt}
\end{figure*}

        Considering the inherent spectral bias of neural networks in synthesizing high-frequency details, we propose to directly extract extra high-frequency embeddings and feed into the decoder. Various methods can be used to extract high-frequency signals from images, such as the discrete cosine transform (DCT) and the discrete wavelet transform (DWT). To obtain disentangled high-frequency information, we employ a simple yet effective Haar wavelet transform \citep{harr} to transform the encoded features from the spatial domain to multiple frequency domains for frequency selection. The Haar wavelet contains four kernels: \(LL^T\), \(LH^T\), \(HL^T\), and \(HH^T\). The transformation is defined as follows:
        \begin{equation}
        L^T=\frac{1}{\sqrt{2}}[1,1],\quad H^{T}=\frac{1}{\sqrt{2}}[-1,1]\
        \end{equation}
        where \(L\) and \(H\) represent the low-pass and high-pass filters, respectively. The transformation decomposes features into four frequency components \(F_{LL}\), \(F_{LH}\), \(F_{HL}\), and \(F_{HH}\). Among these frequency components, \(F_{LL}\) primarily captures low-frequency signals associated with the smooth overall outline of the input image. In contrast, \(F_{LH}\), \(F_{HL}\), and \(F_{HH}\) capture high-frequency signals corresponding to texture details in vertical, horizontal, and diagonal edges, respectively, as illustrated in Fig. \ref{data:dwt} (Left).

        Based on this insight, we propose the Wavelet Frequency Decomposer block (WFD), whose structure is shown in Fig. \ref{data:dwt}. First, we perform the Haar wavelet transform on the input feature \(F \in\mathbb{R}^{C \times H \times W}\) in feature space to separate the frequency components. We then directly sum up the high-frequency components to obtain the high-frequency signal. Then, we employ point convolution operations for feature mapping to obtain low-frequency feature \(F_{L} \in \mathbb{R}^{C \times \frac{H}{2} \times \frac{W}{2}}\) and high-frequency feature \(F_{H} \in \mathbb{R}^{C \times \frac{H}{2} \times \frac{W}{2}}\), respectively. To further leverage the advantages of wavelet multilevel decomposition analysis, we design the wavelet high-frequency encoder using ConvNeXt blocks with strided convolution layers and WFD blocks, as shown in Fig. \ref{data:model}. Specifically, the encoder features undergo frequency separation via the WFD block. The obtained high-frequency features \(F_{H_{i}}\), along with the features \(E_{i-1}\) from the preceding ConvNeXt block, are concatenated and fed into the subsequent ConvNeXt block with strided convolution layer to produce \(E_{i}\) with reduced spatial dimensions. Concurrently, the low-frequency feature \(F_{L_{i}}\) is forwarded to the next stage through skip connections for further decomposition. The output embedding \(E_{i}\) of the \(i\)-th encoder stage can be formulated as:
\begin{equation}
\begin{aligned}
            E_{i} &= ConvNeXt(conv(concat(E_{i-1},F_{H_i})))
           \end{aligned} 
        \end{equation}
        where \(E_i \in\mathbb{R}^{{H_i} \times {W_i} \times {C_i}} ,i = 1,2,3,4 \). 
    \subsection{Frequency-Aware Decoder}
    The task of the decoder is to reconstruct the video frame using compact feature embeddings as inputs, comprising the Harmonic block and the High-Frequency Feature Modulation block, which perform spatial upsampling and fuse features, respectively.
    
\textbf{Harmonic Block.} The decoder typically consisting of stacked upsampling blocks. Each upsampling block generally includes a convolutional layer that expands the channel size, followed by a pixel shuffle layer and an activation layer to increase the spatial resolution of features, as shown in Fig. \ref{data:model} (Right). Previous methods often employ the GELU activation function in these blocks, which can limit the ability of network to capture high-frequency signals in images. \cite{siren} has demonstrated that the activation function is crucial, significantly impacting the convergence and accuracy of the neural network. Drawing inspiration from the real Fourier transform, we proposed the adaptive harmonic activation function to build the Harmonic decoder block. The function is defined as follows:
 \begin{equation}
\sigma(x)=\omega_1\cdot\mathrm{sin}\left(x\right)+\omega_2\cdot\mathrm{cos}\left(x\right)
            \end{equation} 
        where \(\omega_1\) and \(\omega_2\) represent the learnable parameters. 
        This Harmonic block introduces periodic inductive bias while maintaining strong nonlinear capabilities, thereby enhancing the network's ability to capture complex texture patterns and periodic content.
                \begin{figure}[hb]
    \centering
    \scalebox{0.7}{
    \includegraphics[width=1\linewidth]{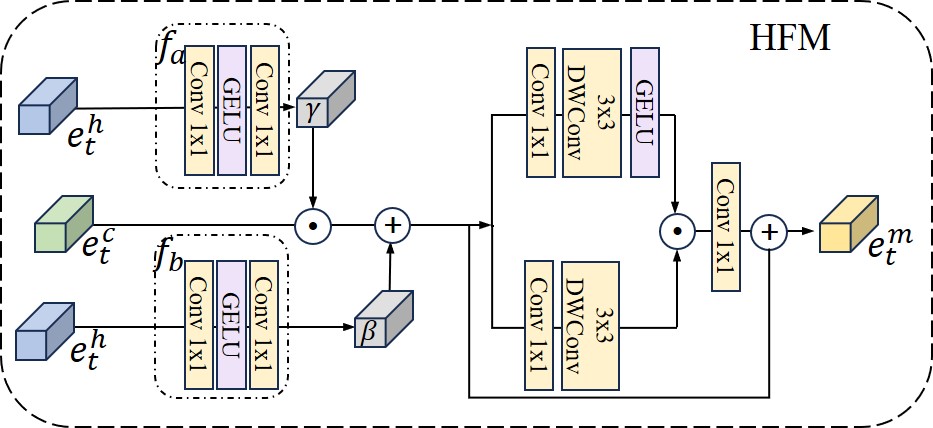}
    }
    \caption{Illustration of the High-Frequency Feature Modulation block (HFM). The HFM learns modulation vectors from high-frequency features to modulate the content features. are then passed through a feed-forward network to further enhance feature representation.}
    \label{data:decoder}
\end{figure}

        \textbf{High-Frequency Feature Modulation Block.} To fully integrate  the obtained content features \(e_t^c\) and high-frequency features \(e_t^h\) from different branches, a straightforward fusion approach like addition or concatenation is inadequate as it may introduce some
        degree of high and low-frequency aliasing.
        In addition, employing a popular attention-based fusion strategy can more precisely combine the features but results in a sharp increase in computational complexity and introduces unnecessary parameter redundancy. To balance the performance and computational cost, we designed a light-weight High-Frequency Feature Modulation (HFM) block, as illustrated in Fig. \ref{data:decoder}. Specifically, the HFM receives high-frequency features as input and predicts the modulation vectors \(\gamma\) and \(\beta\) through two sub-networks \(f_a\) and \(f_b\). Both \(f_a\) and \(f_b\) consist of three convolutional layers, each with a kernel size of 1 × 1. The content features are then modulated by applying the computed \(\gamma\) and \(\beta\), where \(\gamma\) serves as a multiplier and \(\beta\) as an additive bias, allowing the feature maps to be adjusted adaptively. Next, a feed-forward network (FN) \citep{restormer} is employed to further enhance feature representation by improving the modeling of spatial and channel-wise relationships, leading to better performance. This complete operation in HFM can be described as:
\begin{align}
    \gamma,\ \beta &= f_a(e_t^h),\ f_b(e_t^h),\\
    e_t^m &= \mathrm{FN}\big(\gamma \odot e_t^c \oplus \beta\big)
\end{align}

        where \(\odot\) and \(\oplus\) denote the element-wise multiplication and
element-wise addition, respectively. \(e_t^m\) represents the output modified features. This modulation enables the layer to adaptively enhance features based on the high-frequency content, encouraging the decoder to synthesize images with more details and fewer artifacts.
    \subsection{Loss Function}
       Given our goal of reconstructing video frames with fine detail, we opted for a combination of L1 loss with SSIM as a training loss function. The spatial loss \(L_{spa}\) is defined as:
          \begin{equation}
            L_{spa}=\alpha\cdot|\widehat{x}_t-x_t|+(1-\alpha)\cdot(1-\mathrm{ssim}(\widehat{x}_t,x_t))
            \end{equation}
    where \(\alpha\) is a balancing factor, set to 0.7, following the same settings as \citep{nerv}.
    However, relying solely on spatial domain losses may not capture the high-frequency details essential for preserving fine textures. Therefore, we incorporate a frequency loss during training. Specifically, we perform Fast Fourier Transforms (FFTs) on both the reconstructed image \(\widehat{x}_t\) and the ground truth image \(x_t\) to convert them into the frequency domain and compute the L1 loss between these frequency representations. To focus the network on synthesizing challenging high-frequency details, we apply a dynamic weighting strategy inspired by \citep{ffl}. We calculate the spectral weight matrix using the spectral difference \({W} = \left(\left|FFT\big(\widehat{x}_t\big) - FFT(x_t)\right|\right)\). However, directly using this spectral difference can lead to large weight values due to extreme differences in certain frequency components, potentially destabilizing the training process. To mitigate this issue, we apply a logarithmic transformation to mitigating the impact of extreme differences:
    \begin{equation} W = \ln\left( 1 + \left| \text{FFT}\left( \widehat{x}_t \right) - \text{FFT}\left( x_t \right) \right| \right). \end{equation}
    This transformation provides effective supervision of the network's adaptive behavior across different frequency components. The frequency loss \(L_{fre}\) is then defined as:            
        \begin{equation}
        L_{fre}={W}\cdot|FFT(\widehat{x}_t)-FFT(x_t)|
        \end{equation}
    The total loss function is formulated as follows:
        \begin{equation}
            L_{total}=L_{spa}+\mu \cdot L_{fre}
        \end{equation}  
      where \(\mu\) is a hyperparameter used to balance the weight of the frequency loss component and spatial loss component.
\section{Experiments}
    \subsection{Datasets and Implementation Details}
  In this section, we conduct comprehensive experiments on the Big Buck Bunny (Bunny) \citep{bigbuckBunny} and UVG \citep{vug} datasets to evaluate the effectiveness of our method. The Bunny dataset consists of a 720 \(\times\) 1280 resolution video with 132 frames. The UVG dataset consists of seven videos at 1080 \(\times\) 1920 resolution with 600 or 300 frames. In line with previous works, we perform a central crop of 640 \(\times\) 1280 or 960 \(\times\) 1920 on the Bunny and UVG datasets. For our experiments, the downsampling factor \(f\) and \(g\) lists for the content encoder and wavelet high-frequency encoder are set as follows: [5, 4, 4, 2, 2] and [2, 2, 2, 2, 2] for the Bunny, [5, 4, 4, 3, 2] and [3, 2, 2, 2, 2] for the UVG, respectively. Similarly, the upsampling factors of the decoder are set to [5, 4, 4, 2, 2] for the Bunny and [5, 4, 4, 3, 2] for the UVG. We set the size of the content embedding as \(16\times2\times4\) and the high-frequency embedding to \(2 \times 10 \times 20\). We employ peak-signal-to-noise ratio (PSNR) and multi-scale structural similarity index (MS-SSIM) to evaluate the reconstruction quality and assess the video compression performance by measuring the bits per pixel (bpp). During training, we use the Adan optimizer \citep{adan} with cosine learning rate decay. The learning rate is set as 3e-3 with 10\% of the total epochs for warm-up. The batch size is set as 1. All experiments are implemented on PyTorch \citep{pytorch} framework on an NVIDIA GeForce RTX 4090 GPU.

    \subsection{Main Results}
      \subsubsection{Video Representation Result}
        We conduct a comparative analysis of our method against NeRV \citep{nerv}, E-NeRV \citep{E-nerv}, and HNeRV \citep{hnerv}. We trained all methods for 300 epochs using the default hyperparameters and loss functions of each method. We adjust the number of channels in the decoder to obtain representation models of different sizes. Table \ref{table:data1} shows the regression performance of various methods on the Bunny dataset at different model sizes. The comparison results illustrate that our method outperforms other methods across all sizes. Table \ref{table:data2} presents the performance across varying training epochs, further emphasizing the robustness of our approach.

\begin{minipage}{0.9\textwidth}
        \begin{minipage}[t]{0.52\textwidth}
            \begin{table}[H]
                \centering
                \scalebox{0.68}{
            \begin{tabular}{@{}c|cccc@{}}
            \hline
            Size   & 0.35M & 0.75M & 1.5M  & 3M    \\ 
            \hline
            NeRV   & 26.99 & 28.46 & 30.87 & 33.21 \\
            E-NeRV & 27.84 & 30.95 & 32.09 & 36.72 \\
            HNeRV & 30.15 & 32.81 & 35.19 & 37.43 \\ \hline
            Ours   & \textbf{31.71} & \textbf{34.67} & \textbf{37.64} & \textbf{40.33} \\ \hline
            \end{tabular}
            }
                \caption{PSNR with varying model size on Bunny. Bold means best results.}
                \label{table:data1}
                \vspace{3pt}
            \end{table}
        \end{minipage}
        \quad
        \begin{minipage}[t]{0.45\textwidth}
            \begin{table}[H]
                \centering
               \scalebox{0.68}{
            \begin{tabular}{@{}c|ccc@{}}
            \hline
            Epoch   & 300          & 600          & 1200\\   \hline
            NeRV & 33.21 & 34.47 & 35.07 \\
            E-NeRV & 36.72 & 38.20 & 39.48 \\
            HNeRV & 37.43 & 39.36 & 40.02 \\
            \hline
            Ours     & \textbf{40.33}& \textbf{40.80}& \textbf{41.18}\\ 
            \hline
            \end{tabular}
            }
            \caption{PSNR with varying epochs on Bunny. Bold means best results.}
            \label{table:data2}
                \vspace{3pt}
            \end{table}
        \end{minipage}
         \vspace{5pt}
    \end{minipage}

\begin{table}[htb]
\centering
\renewcommand{\arraystretch}{1.2}
\scalebox{0.5}{
\begin{tabular}{@{}c|c|cccccccc@{}}
\hline
\makebox[0.1\textwidth][c]{Resolution} & Method & Beauty & Honey. & Bosph. & Yacht. & Ready. & Jockey & Shake. & Average \\
\hline
\multirow{4}{*}{960\(\times\)1920} & NeRV & 33.25/0.8886 & 37.26/0.9794 & 33.22/0.9305 & 28.03/0.8726 & 24.84/0.8310 & 31.74/0.8874 & 33.08/0.9325 & 31.63/0.9031 \\
& E-NeRV & 33.53/0.8958 & 39.04/0.9845 & 33.81/0.9442 & 27.74/0.8951 & 24.09/0.8515 & 29.35/0.8805 & 34.54/0.9467 & 31.73/0.9140 \\
& HNeRV & 33.58/0.8941 & 38.96/0.9844 & 34.73/0.9451 & 29.26/0.8907 & 25.74/0.8420 & 32.04/0.8802 & 34.57/0.9450 & 32.69/0.9116 \\
& Ours & \textbf{33.92}/\textbf{0.8996} & \textbf{39.44}/\textbf{0.9850} & \textbf{35.53}/\textbf{0.9554} & \textbf{29.89}/\textbf{0.9084} & \textbf{27.13}/\textbf{0.8910} & \textbf{33.71}/\textbf{0.9157} & \textbf{35.29}/\textbf{0.9537} & \textbf{33.56}/\textbf{0.9298} \\
\hline
\multirow{4}{*}{480\(\times\)960} & NeRV & 32.38/0.9346 & 36.64/0.9912 & 32.95/0.9577 & 28.07/0.9183 & 24.55/0.8884 & 31.33/0.9154 & 32.74/0.9603 & 31.24/0.9380 \\
& E-NeRV & 32.59/0.9399 & 38.47/0.9936 & 33.72/0.9705 & 27.86/0.9393 & 24.05/0.9069 & 28.98/0.9084 & 34.06/0.9715 & 31.39/0.9472 \\
& HNeRV & 32.81/0.9341 & 38.52/0.9936 & 34.58/0.9703 & 29.24/0.9354 & 25.73/0.9112 & 32.04/0.9151 & 34.34/0.9698 & 32.47/0.9470 \\
& Ours & \textbf{33.01}/\textbf{0.9430} & \textbf{38.76}/\textbf{0.9940} & \textbf{35.52}/\textbf{0.9770} & \textbf{30.03}/\textbf{0.9467} & \textbf{27.53}/\textbf{0.9434} & \textbf{33.60}/\textbf{0.9429} & \textbf{34.84}/\textbf{0.9754} & \textbf{33.33}/\textbf{0.9603}\\ 
\hline
\end{tabular}
}
\caption{Video representation results in terms of PSNR and MS-SSIM on UVG at resolution 960\(\times\)1920 and 480\(\times\)960 with 3M model size, respectively. Bold means best results.}
\label{table:data3}
\end{table}

\begin{figure*}[htb]
	\centering
	\begin{minipage}{0.24\linewidth}
		\centering
		\includegraphics[width=1\linewidth]{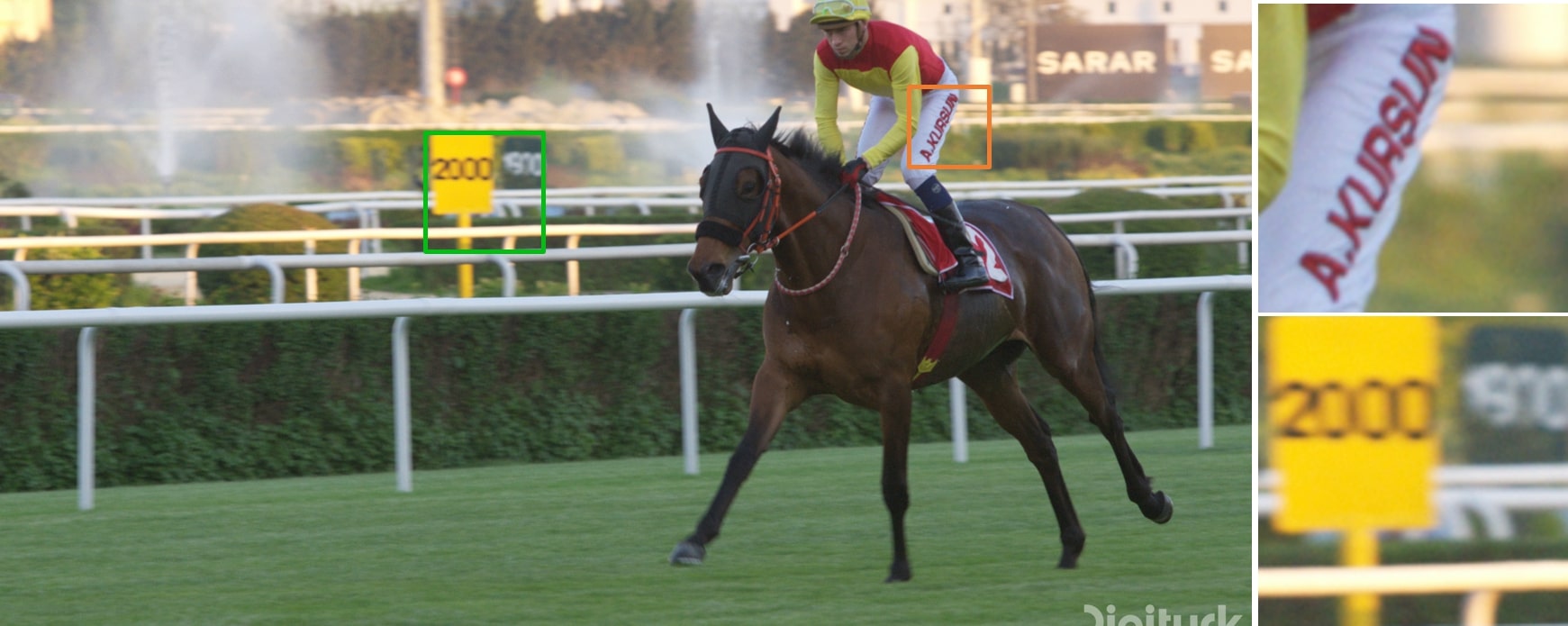}
	\end{minipage}
 \begin{minipage}{0.24\linewidth}
		\centering
		\includegraphics[width=1\linewidth]{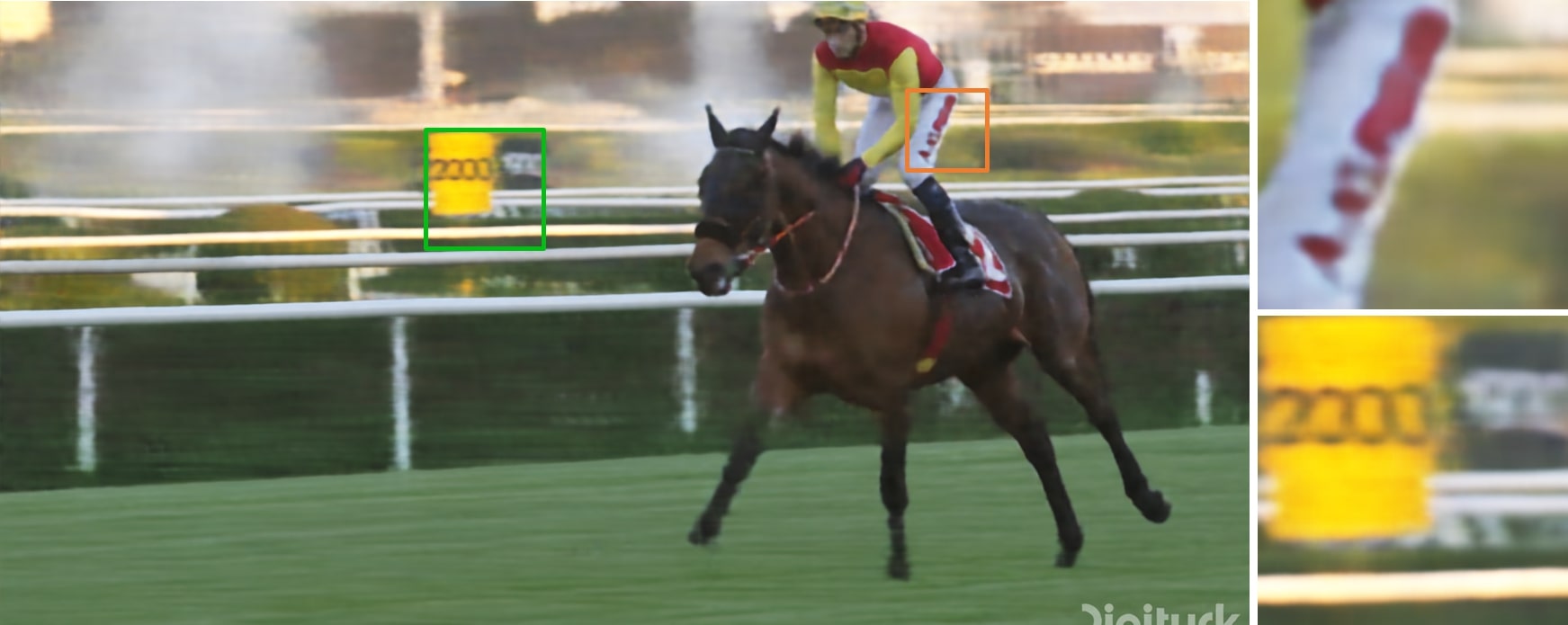}
	\end{minipage}
	\begin{minipage}{0.24\linewidth}
		\centering
		\includegraphics[width=1\linewidth]{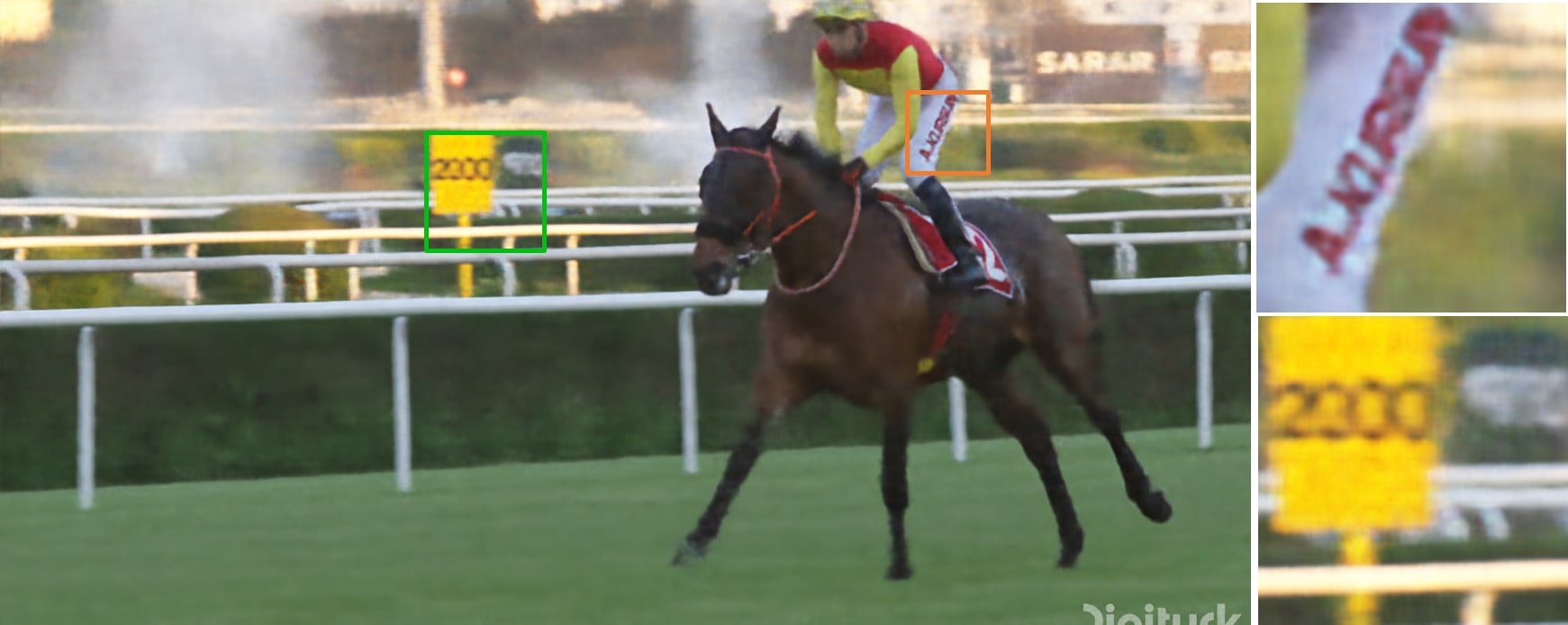}
	\end{minipage}
     \begin{minipage}{0.24\linewidth}
        \centering
        \includegraphics[width=1\linewidth]{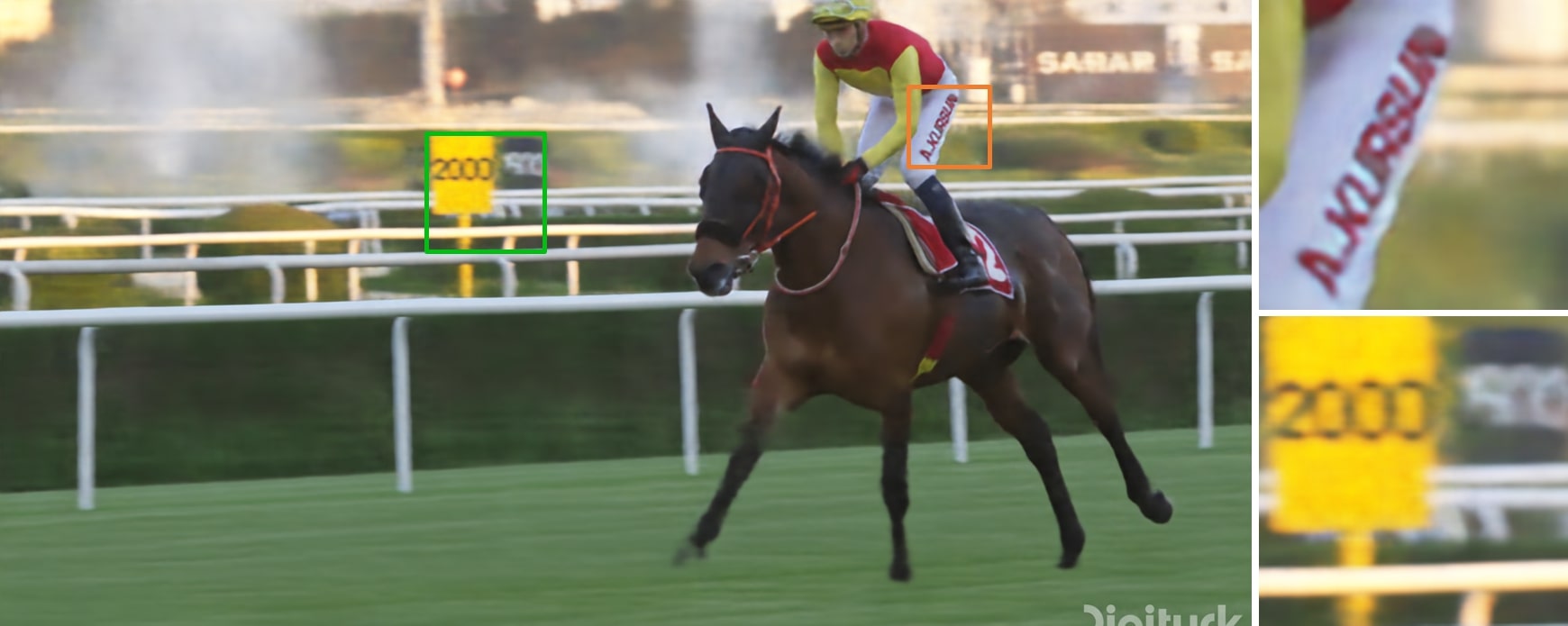}
    \end{minipage}
	\begin{minipage}{0.24\linewidth} 
		\centering
		\includegraphics[width=1\linewidth]{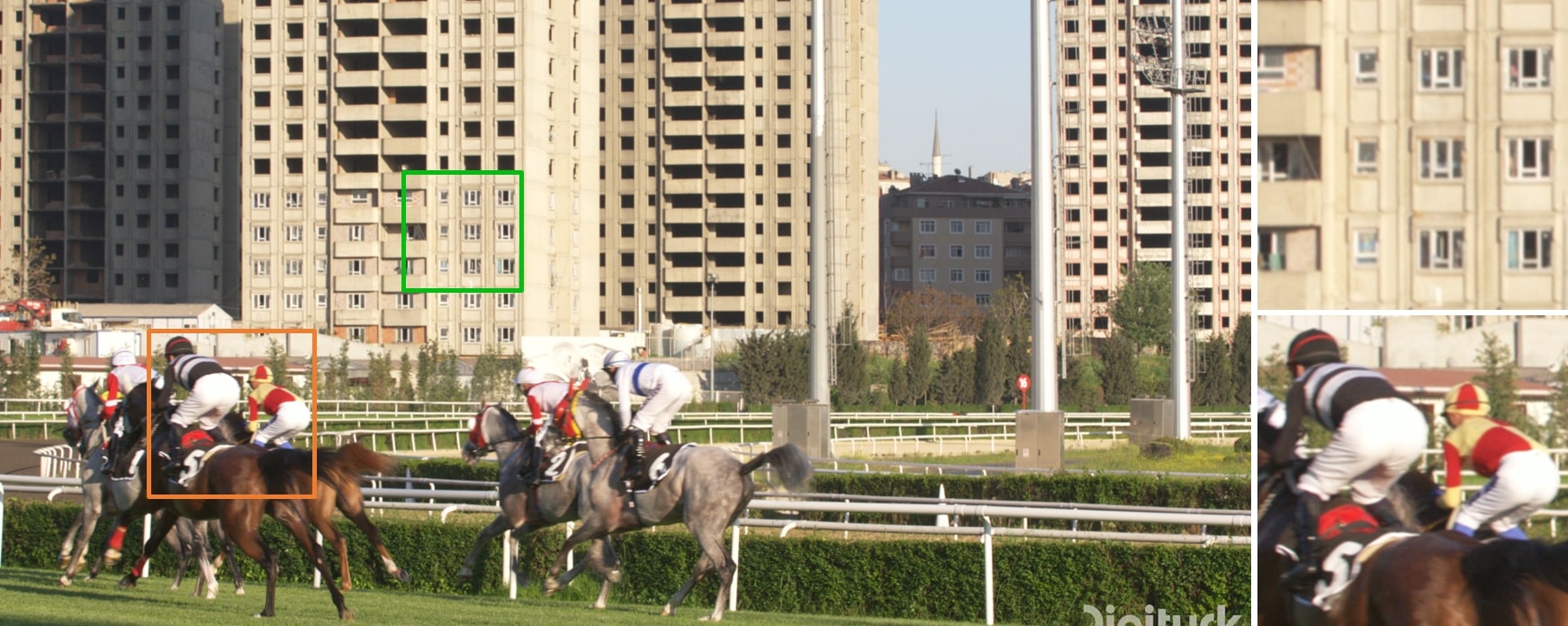}
	\end{minipage}
    \begin{minipage}{0.24\linewidth} 
		\centering
		\includegraphics[width=1\linewidth]{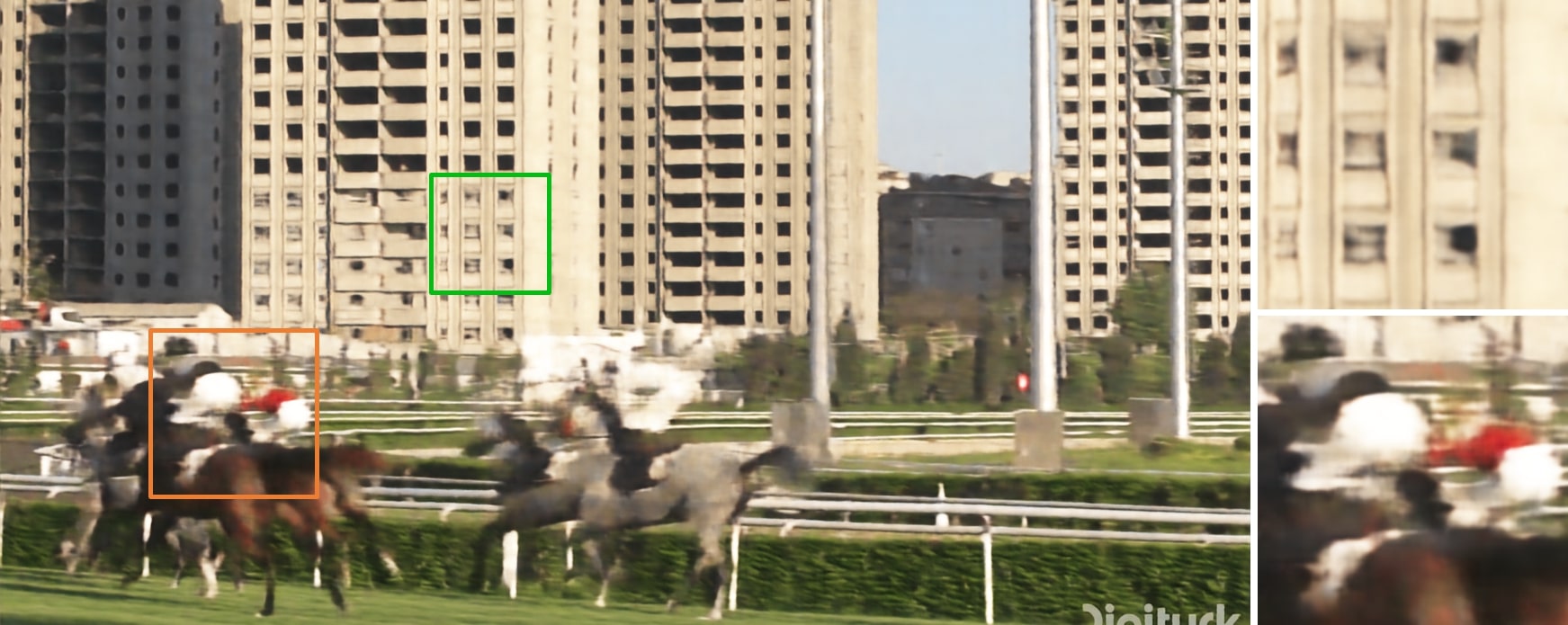}
	\end{minipage}
	\begin{minipage}{0.24\linewidth}
		\centering
		\includegraphics[width=1\linewidth]{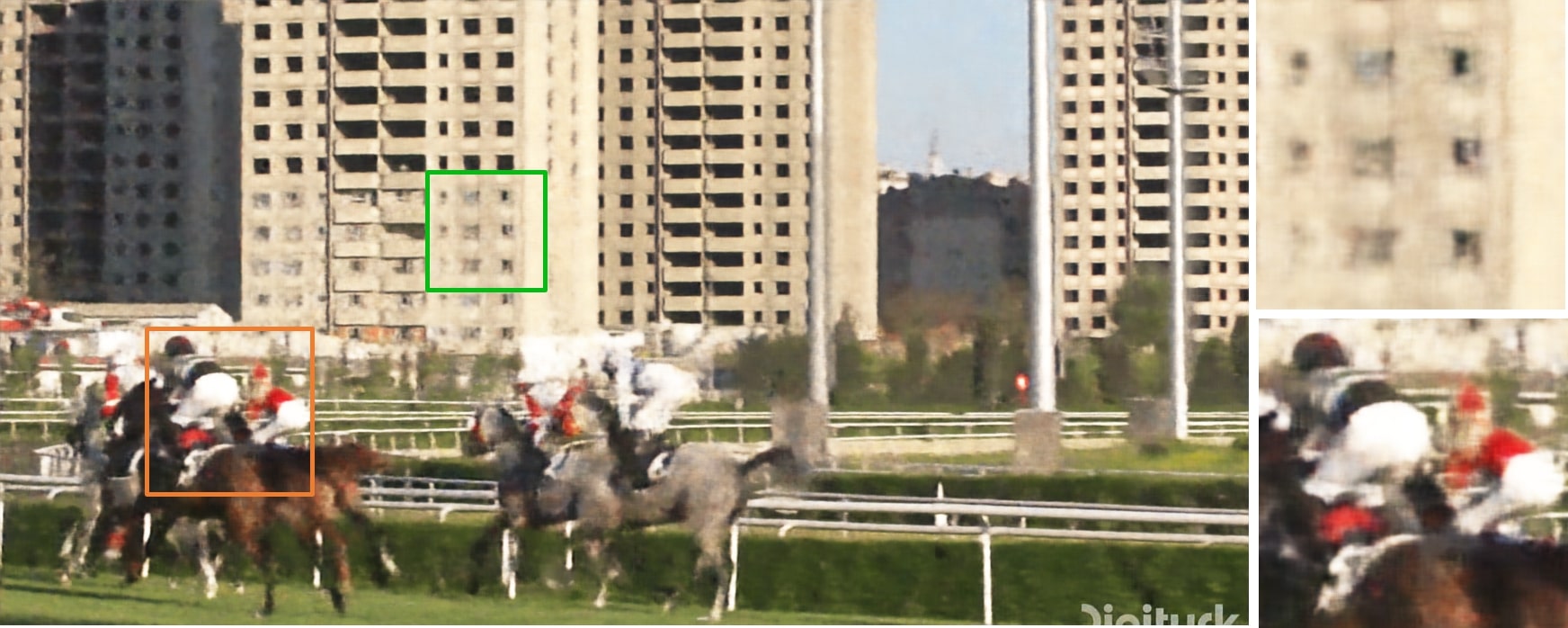}
	\end{minipage}
     \begin{minipage}{0.24\linewidth}
        \centering
        \includegraphics[width=1\linewidth]{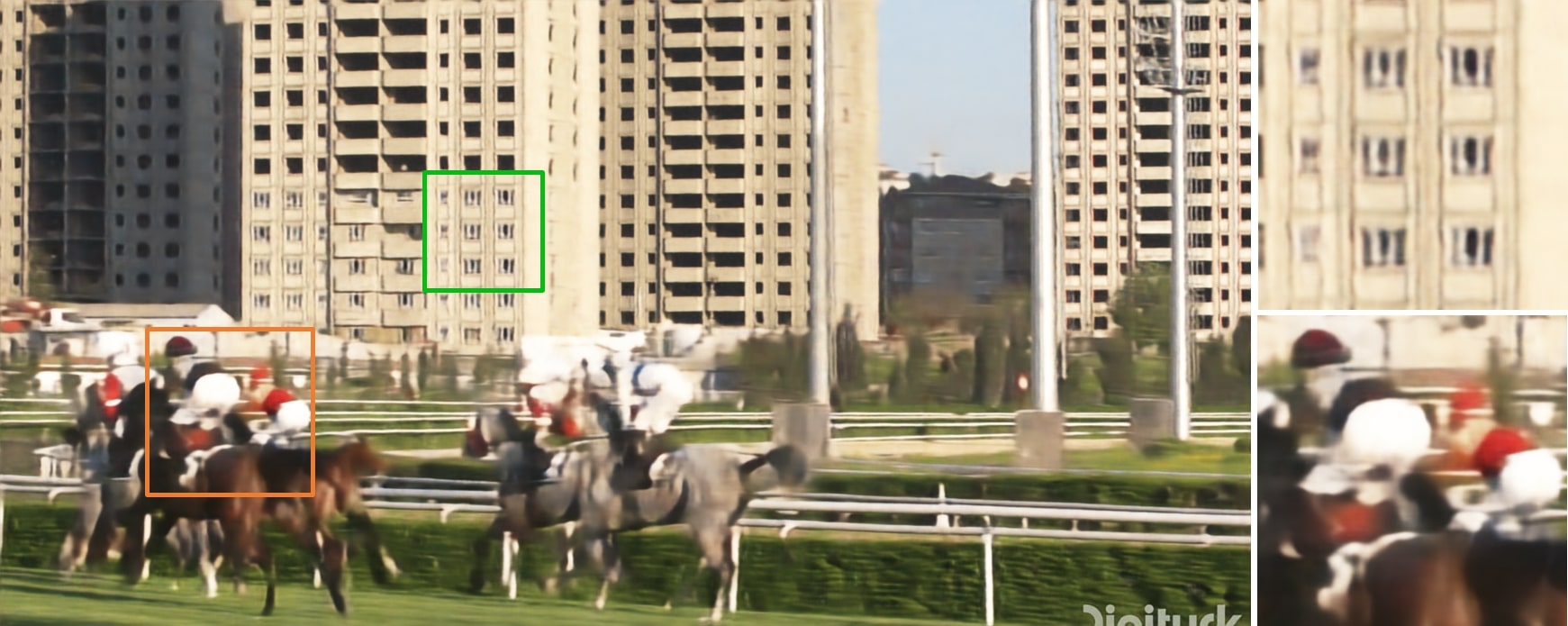}
    \end{minipage}
	\begin{minipage}{0.24\linewidth}
		\centering
		\includegraphics[width=1\linewidth]{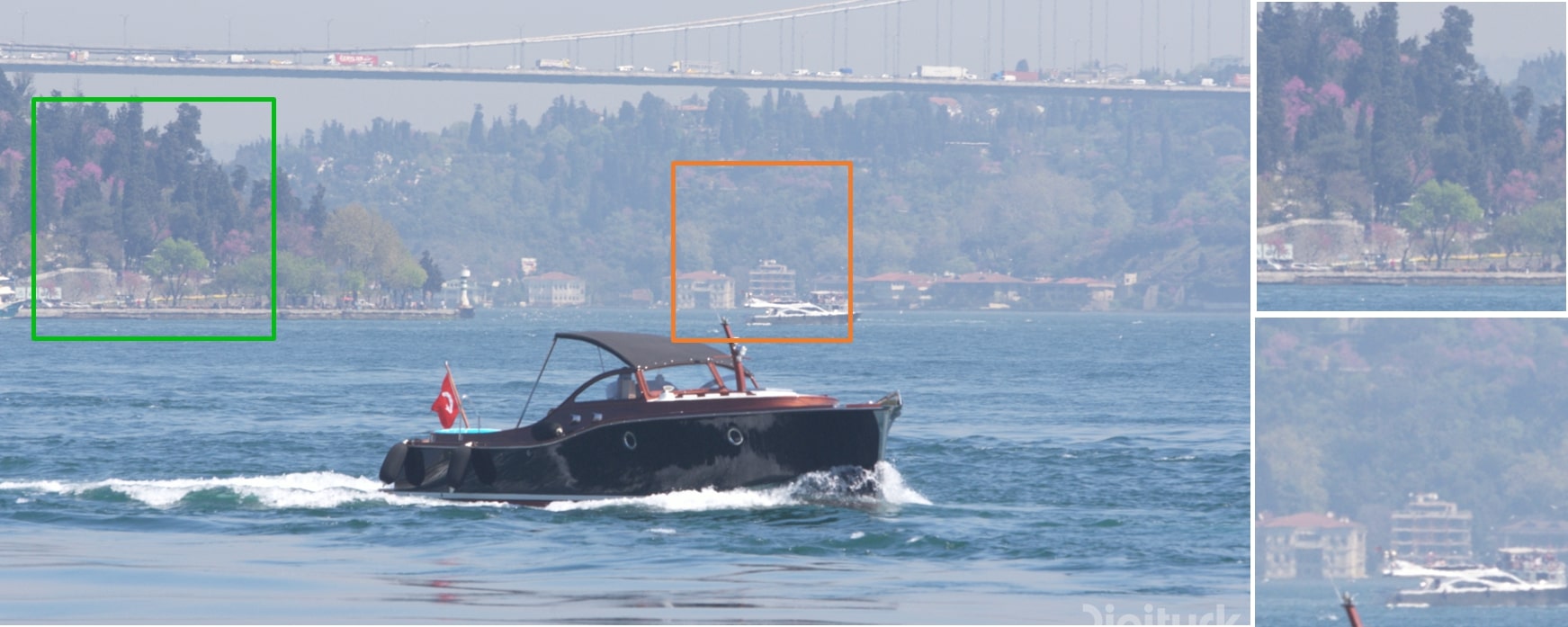}
	\end{minipage}
  \begin{minipage}{0.24\linewidth}
		\centering
		\includegraphics[width=1\linewidth]{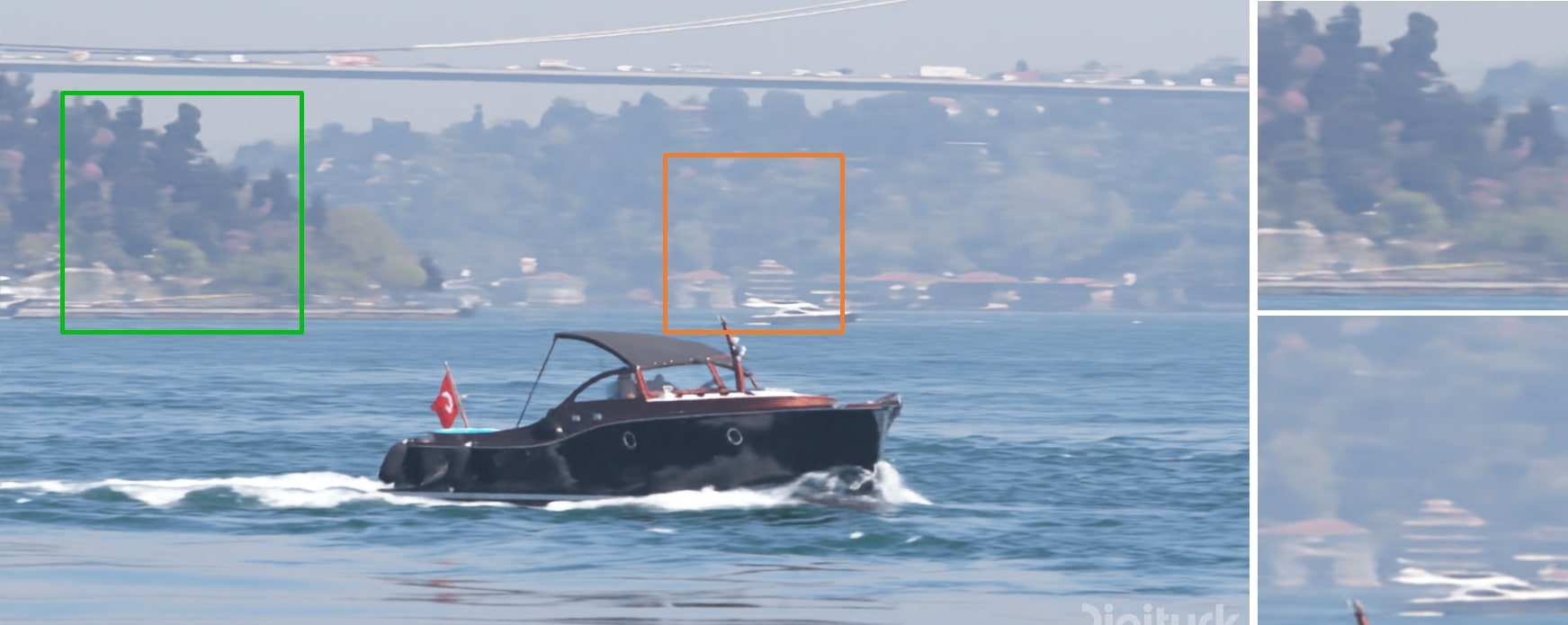}
	\end{minipage}
	\begin{minipage}{0.24\linewidth}
		\centering
		\includegraphics[width=1\linewidth]{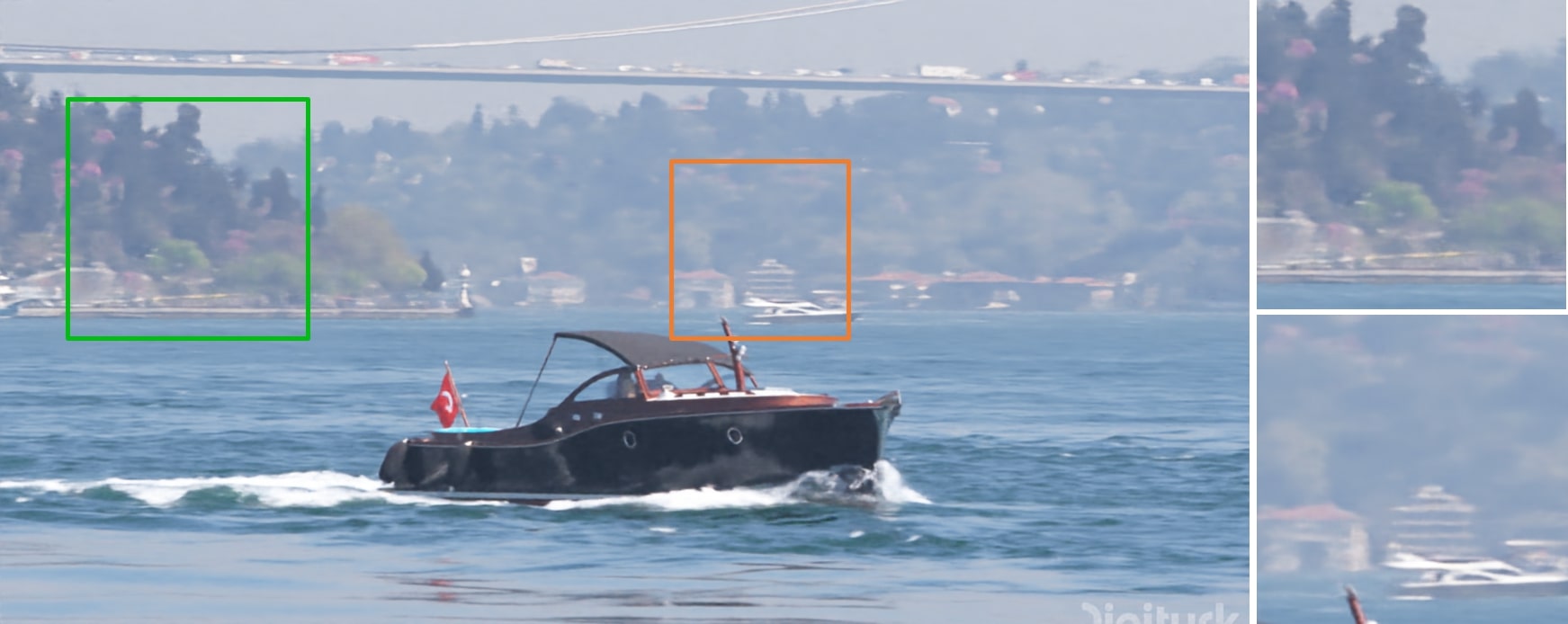}
	\end{minipage}
     \begin{minipage}{0.24\linewidth}
        \centering
        \includegraphics[width=1\linewidth]{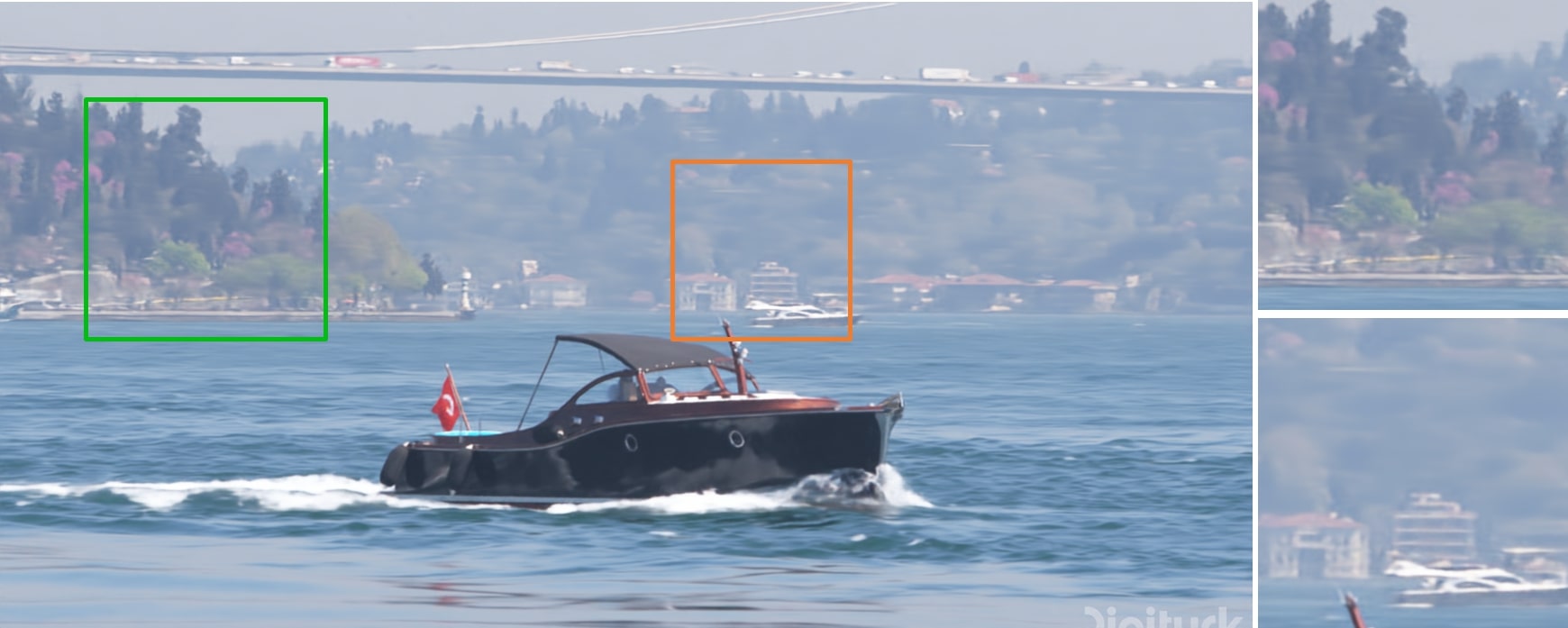}
    \end{minipage}
    \caption{Video reconstruction visualization of UVG dataset. The first column is the ground truth, the second and third columns are the reconstruction result of E-NeRV and HNeRV, and the fourth column is the reconstruction result of our method. Our approach demonstrates superior performance in preserving texture structure.}
	\label{data:Visualization}
    \end{figure*}

        The first row of Table \ref{table:data3} lists the video regression results on the UVG dataset. Our method outperformed others overall, achieving an average improvement of +0.87 dB in PSNR and +0.0182 in MS-SSIM. It is worth noting that prominent improvements can be observed on \(ReadySteadyGo\) and \(Jackey\) sequences because of their complex texture details, achieving an improvement of +1.39 dB and +1.67 dB in PSNR, respectively. Conversely, on video with simpler textures like \(Bee\), the improvements were less pronounced. 
        We implement performance validation on the downsampled version of UVG (480\(\times\)960), and the results demonstrate the superiority and generalizability of our method. We further visualize the reconstruction results of the different methods, as shown in Fig. \ref{data:Visualization}. Our method demonstrates an advantage in maintaining image details, such as the text on the leg of the jockey in the first row and the structure of the window in the second row. Benefiting from the innovative network structure design, the performance of our method is significantly improved compared to other methods in video regression tasks.
          
      \begin{figure}[hb]
        \centering
        \scalebox{0.99}{
    	
    	\begin{minipage}{0.49\linewidth}
    		\centering
    		\includegraphics[width=1\linewidth]{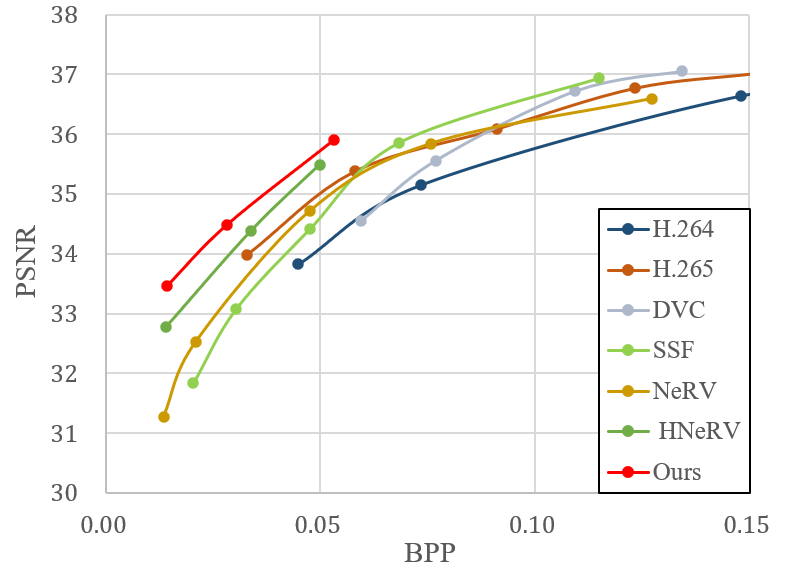}
    	\end{minipage}
    	\begin{minipage}{0.49\linewidth}
    		\centering
    		\includegraphics[width=1\linewidth]{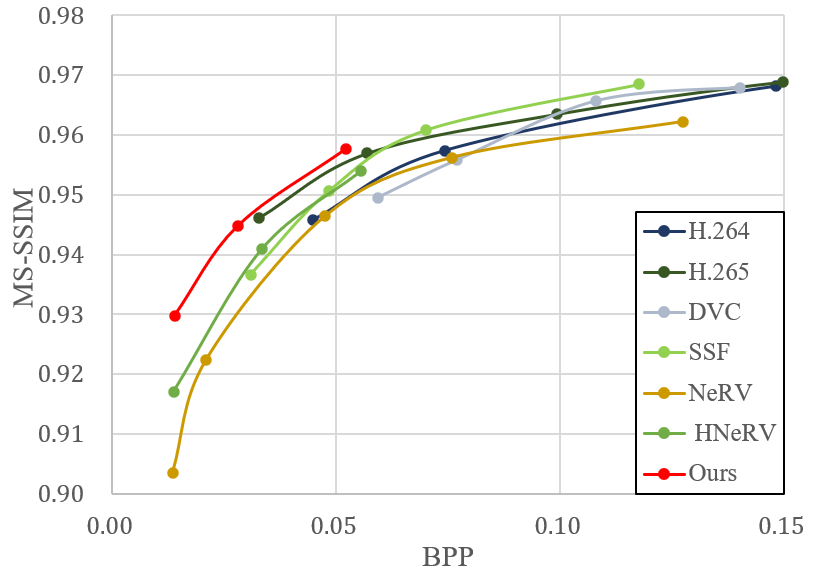}
    		
    	\end{minipage}
     }
            \caption{The rate-distortion curve on UVG in terms of PSNR and MS-SSIM.}
		\label{fig:data1}
        \end{figure}

    \subsubsection{Video Compression Performance}
         After fully training the model, we apply global unstructured pruning using the Layer Adaptive Magnitude Pruning (LAMP) score \citep{layer} to identify and remove less significant parameters, effectively reducing the model's complexity without significantly impacting performance. We then implement 8-bit quantization to reduce the precision of model weights and Huffman entropy coding \citep{huffman} to further compress the model size. We trained three models of different sizes and compressed them to obtain rate-distortion curves, which illustrate the trade-off between compression rate and reconstruction quality. We compare the compression performance of our method with NeRV, HNeRV, and various neural video codecs, including DVC \citep{dvc}, scale-space-flow model(SSF) \citep{ssf}, as well as traditional codecs like H.264 \citep{h264} and H.265 \citep{h265}. Fig. \ref{fig:data1} presents the rate-distortion curves evaluated on the UVG dataset. The results indicate that our method not only outperforms other INR models but also surpasses both end-to-end neural video codecs and traditional video codecs. This highlights the effectiveness of our compression strategy and framework design.

\begin{minipage}{1\textwidth}
    \begin{minipage}[t]{0.45\textwidth}
        \begin{table}[H]
            \centering
            \scalebox{0.75}{
            \begin{tabular}{@{}l|cc@{}}
            \hline
            \multicolumn{1}{c}{Variant} & PSNR  & MS-SSIM \\ \hline
            Ours                          & \textbf{40.33} & \textbf{0.9929}  \\ \hline
            -w/o High-frequency(V1) & 39.08 & 0.9914  \\
            -w/ Content encoder(V2)       & 40.18 & 0.9927  \\
            \hline
            \end{tabular}
            }
            \caption{Ablation on high-frequency encoder.}
            \label{table:part1}
        \end{table}
    \end{minipage}
    \quad
    \begin{minipage}[t]{0.45\textwidth}
        \begin{table}[H]
            \centering
            \scalebox{0.75}{
            \begin{tabular}{@{}l|cc@{}}
            \hline
            \multicolumn{1}{c}{Variant} & PSNR  & MS-SSIM \\ \hline
            -w/ Concat(V3)                & 39.28 & 0.9914 \\
            -w/ Add(V4)                   & 39.36 & 0.9915 \\
            -w/ Inter-attention (V5)      & 39.83 & 0.9923 \\
            \hline
            \end{tabular}
            }
            \caption{Ablation on fusion strategies.}
            \label{table:part2}
        \end{table}
    \end{minipage}
    \vspace{10pt}
    \begin{minipage}[t]{0.45\textwidth}
        \begin{table}[H]
            \centering
            \scalebox{0.75}{
            \begin{tabular}{@{}l|cc@{}}
            \hline
            \multicolumn{1}{c}{Variant} & PSNR  & MS-SSIM \\ \hline
            -w/ GELU(V6)                 & 40.03 & 0.9925  \\
            -w/ Sinusoidal(V7)           & 40.12 & 0.9927  \\
            \hline
            \end{tabular}
            }
            \caption{Ablation on activation function.}
            \label{table:part3}
        \end{table}
    \end{minipage}
    \quad\quad\quad
    \begin{minipage}[t]{0.45\textwidth}

        \begin{table}[H]
            \centering
            \scalebox{0.75}{
            \begin{tabular}{@{}l|cc@{}}
            \hline
            \multicolumn{1}{c}{Variant} & PSNR  & MS-SSIM \\ \hline
            -w/ L1+SSIM(V8)              & 39.94 & 0.9929  \\ 
            -w/ L2(V9)                   & 39.88 & 0.9918  \\  
            -w/o Log(V10)                & 40.23 & 0.9929  \\
            \hline
            \end{tabular}
            }
            \caption{Ablation study on loss functions.}
            \label{table:part4}
        \end{table}

    \end{minipage}

\end{minipage}

    \subsubsection{Ablation Study}
    In this section, we conduct ablation studies on the Bunny dataset to verify the contribution of various components in our method. We generated multiple variants of the original model and trained them over 300 epochs to evaluate their performance.

\begin{figure}[htb]
    \centering
    \scalebox{0.95}{
    \includegraphics[width=0.49\linewidth]{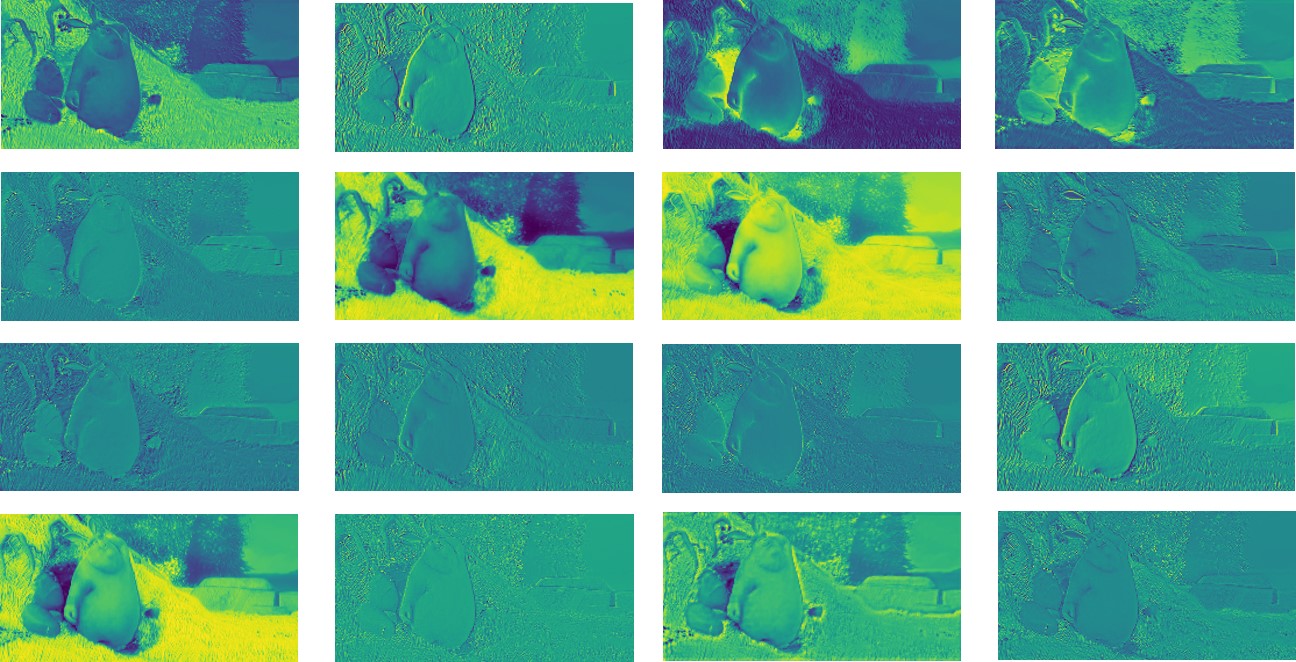}
    \quad 
    \includegraphics[width=0.49\linewidth]{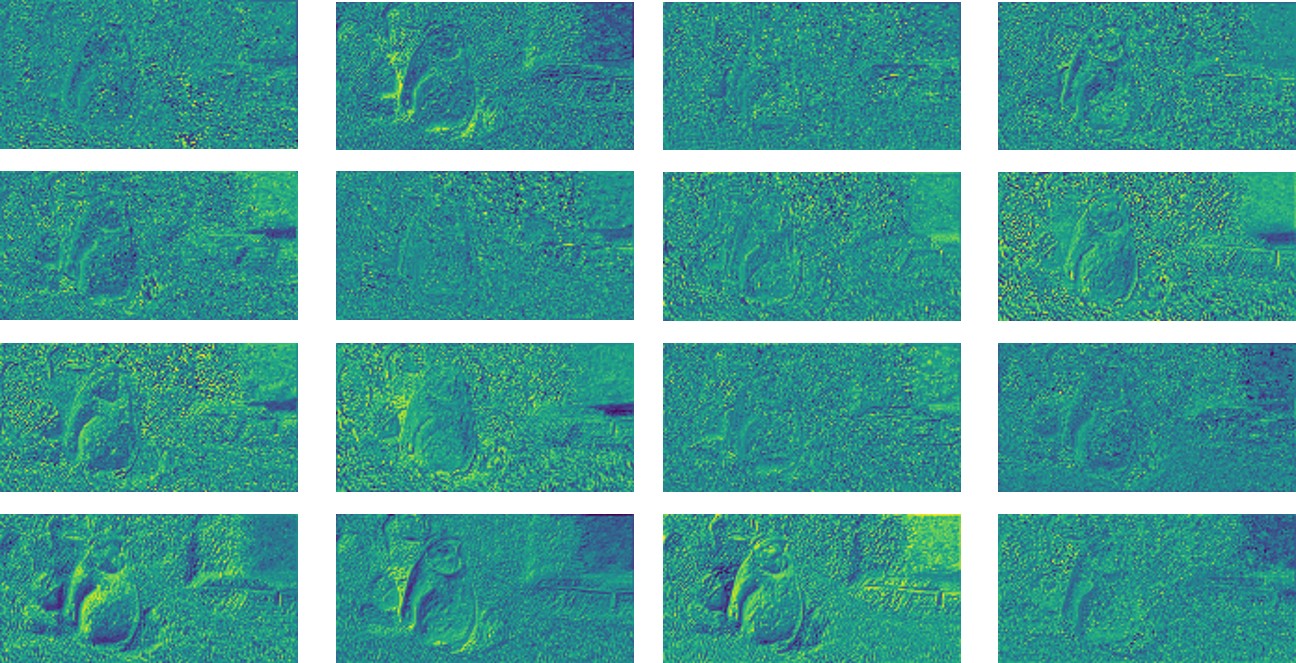}
    }
    \caption{Feature map visualization results of the content encoder (Left) and wavelet high-frequency encoder (Right). We select the first 16-channel feature maps from the second stage of the 114-frame generation of the Bunny video.}
    \label{data:fig6} 
\end{figure}

    \textbf{High-frequency encoder.} To evaluate the effectiveness of High-frequency encoder, we conducted two experiments as shown in Table \ref{table:part1}. In the first variant (V1), we removed the high-frequency embedding from our model while keeping the number of parameters in V1 the same as in the original model, where the average PSNR drops by 1.25 dB. In the second variant (V2), we replaced the wavelet high-frequency encoder with the content encoder, resulting in an average PSNR decrease of 0.15 dB. As shown in Fig. \ref{data:fig6}, we further visualize the intermediate feature maps of the two encoders, showing that the content encoder captures low-frequency information such as image structure, while the wavelet high-frequency encoder captures more complex texture information. This observation highlights the capability of our wavelet high-frequency encoder to independently extract the desired high-frequency features effectively.

   \textbf{High-frequency Embedding Size.} The total size of the video representation is influenced by both the embedding size and the model parameters. Increasing the embedding size typically reduces the number of model parameters, which can impact performance. To identify the optimal balance between embedding size and model efficiency, we conducted an ablation study. As shown in Table \ref{table:embeddingsize}, we observed that performance improves as the embedding size increases, but excessive size will limit the number of trainable parameters available for the decoder, leading to performance degradation. Specifically, when the spatial dimensions are set to $10 \times 20$ with 2 channels, the model achieves the best trade-off between compactness and performance.

\begin{minipage}[hb]{0.3\linewidth}  
    \begin{table}[H]
        \centering
        \scalebox{0.7}{
            \begin{tabular}{@{}l|cc@{}}
            \hline
            \multicolumn{1}{c}{Size} & \multicolumn{1}{c}{PSNR} & \multicolumn{1}{c}{MS-SSIM} \\ \hline
            $4 \times 8 \times 2$  & 40.07 & 0.9926  \\
            \textbf{$10 \times 20 \times 1$} & 40.16 & 0.9925  \\
                    \textbf{$10 \times 20 \times 2$} & \textbf{40.33} & \textbf{0.9929}  \\
            $10 \times 20 \times 4$ & 40.24 & 0.9928  \\
            $20 \times 40 \times 2$ & 40.11 & 0.9925  \\
            
            \hline
            \end{tabular}
        }
        \caption{Ablation on High-frequency embedding size.}
        \label{table:embeddingsize}
    \end{table}
        \vspace{5pt}
\end{minipage}
\hfill
\begin{minipage}[htb]{0.3\linewidth}  
    \begin{table}[H]
        \centering
        \scalebox{0.7}{
            \begin{tabular}{@{}l|cc@{}}
            \hline
            \multicolumn{1}{c}{Wavelet} & PSNR  & MS-SSIM \\ \hline
            haar                        & \textbf{40.33} & \textbf{0.9929}  \\  
            db2                         & 40.32 & 0.9929  \\ 
            db4                         & 40.20 & 0.9927  \\  
            db8                         & 40.19 & 0.9927  \\
            sym4                        & 40.25 & 0.9928  \\
            \hline
            \end{tabular}
        }
        \caption{Ablation on different wavelet type.}
        \label{table:wavelet}
    \end{table}
        \vspace{5pt}
\end{minipage}
\hfill
\begin{minipage}[htb]{0.3\linewidth}  
    \begin{table}[H]
        \centering
        \scalebox{0.7}{
            \begin{tabular}{@{}l|cc@{}}
            \hline
            \multicolumn{1}{c}{$\mu$} & PSNR  & SSIM \\ \hline
            1  & 40.12 & 0.9925 \\
            10  & 40.27 & 0.9927 \\
             100  & \textbf{40.33} & \textbf{0.9929} \\
            1000 & 40.05 & 0.9920 \\
            10000 & 40.00 & 0.9918 \\
            \hline
            \end{tabular}
        }
        \caption{ Ablation on different balance factors $\mu$.}
        \label{table:mu}
    \end{table}
    \vspace{5pt}
\end{minipage}

    \textbf{Wavelet type.} To evaluate the effectiveness of different wavelet types in extracting high-frequency features, we conducted experiments comparing several common wavelets, including the Haar wavelet, Daubechies wavelets (db2, db4, db8), and Symlets wavelet (sym4). The results presented in Table \ref{table:wavelet} show that the performance differences among the wavelets are minimal, with the Haar and db2 wavelets performing slightly better. Given the Haar wavelet's computational simplicity and efficiency, we ultimately selected it as the basis for the Wavelet Frequency Decomposer (WFD) module.

    \textbf{Fusion strategy.} We conducted experiments to investigate the effectiveness of HFM. We replaced our proposed HFM with different fusion strategies. In variant (V3), we concatenated the two features along the channel dimension and then processed them further through a convolution layer. In variant (V4), we added the high- and low-frequency features element-by-element. In variant (V5), we performed inter-attention based on the channel dimension to achieve interactive fusion. As shown in Table \ref{table:part2}, the result shows our proposed method achieves the best performance, demonstrating the effectiveness of HFM.

    \textbf{Harmonic block.} To verify the effectiveness of the Harmonic block, we replace its harmonic activation function with the GELU activation function and the sinusoidal activation function \citep{siren} for variant (V6) and variant (V7). The result in Table \ref{table:part3} indicates that the adaptive harmonic activation facilitates the reconstruction of the details of the image by introducing periodic waveform features into the neural network.

    \textbf{Loss function.} 
In Variant (V8), we removed the frequency loss component, using only the spatial loss. This led to a decrease in reconstruction quality, with a 0.39 dB drop in PSNR compared to our full loss function. In Variant (V9), we replaced our loss function with L2 loss alone, resulting in an even greater performance reduction, with a 0.45 dB decrease in PSNR. In Variant (V10), we evaluated the impact of the logarithmic function in the spectral weighting matrix by excluding it from the loss function. The results in Table \ref{table:part4} highlight the effectiveness of our loss function design. Additionally, the parameter $\mu$ governs the balance between frequency loss and pixel-level loss. A large value of $\mu$ leads the model to overly prioritize high-frequency information, while a small $\mu$ value may result in insufficient attention to high-frequency details, negatively impacting the textural quality essential for perceptual fidelity. Experimental results, as shown in Table \ref{table:mu}, demonstrate that setting $\mu$ to 100 yields the best performance, striking an optimal balance between detail preservation and reconstruction quality.

\section{Conclusion}
    In this paper, we introduce a High-Frequency Enhanced Hybrid Neural Representation Network to enhance the video fitting process. By extracting and leveraging high-frequency information and introducing refined Harmonic decoder block, we improve the capacity of the network to accurately regress intricate details in the reconstructed videos. With a Dynamic Weighted Frequency Loss, our approach demonstrates the superiority in reconstructing fine image details on Bunny and UVG datasets. A limitation of our work is that our method primarily focuses on utilizing high-frequency information in the spatial dimension, and the high-frequency information in the temporal dimension has not been fully explored. This may lead to less optimal video reconstruction quality, particularly in scenes with rapidly changing motion. In future work, we aim to explore how to integrate temporal information, such as optical flow, into our model to improve both reconstruction quality and efficiency.
	

	
	
	
	\bibliography{ref_Revised}
	
\end{document}